\pdfoutput=1
\documentclass[11pt]{article}

\usepackage[]{acl}

\usepackage{times}
\usepackage{latexsym}

\usepackage[T1]{fontenc}
\usepackage[utf8]{inputenc}

\usepackage{microtype}
\usepackage{graphicx}
\graphicspath{{image/}}
\usepackage{pdfpages}
\usepackage{url}
\usepackage{hyperref}
\usepackage{xcolor}
\usepackage{epsfig}
\usepackage{adjustbox}
\usepackage{amsfonts}
\usepackage{amsmath}
\usepackage{amssymb}
\usepackage{booktabs} 
\usepackage{comment}
\usepackage{multirow}
\usepackage{caption}
\usepackage{subcaption}
\usepackage{textcomp}
\usepackage{comment}
\usepackage{relsize}
\usepackage{stmaryrd}
\usepackage{bbm}
\usepackage{rotating}

\title{Mind the Gap! Injecting Commonsense Knowledge for Abstractive Dialogue Summarization}

\author{~Seungone Kim\textsuperscript{\rm 1}\thanks{~~Equal contribution}~~~~~~~~~~~~~ Se June Joo\textsuperscript{\rm 1}$\textbf{}^{*}$~~~~~~~~~~ Hyungjoo Chae\textsuperscript{\rm 1,2}$\textbf{}^{*}$ \\\textbf{Chaehyeong Kim}\textsuperscript{\rm 1}$\textbf{}^{*}$~~~~ \textbf{Seung-won Hwang}\textsuperscript{\rm 3}~~~~~  \textbf{Jinyoung Yeo}\textsuperscript{\rm 1,2}\thanks{~~Corresponding author}\\
  Yonsei University\textsuperscript{\rm 1} ~~~
  Tutoring at Market Designers\textsuperscript{\rm 2} ~~~
  Seoul National University\textsuperscript{\rm 3}\\
  \texttt{\{louisdebroglie, sr7418, mapoout, cheris8, jinyeo\}@yonsei.ac.kr}\\
  \texttt{seungwonh@snu.ac.kr}\\}
  
\begin{document}
\maketitle

\newcommand{\mcal}[1]{{\cal{#1}}}
\newcommand{\calA}{\mbox{${\cal A}$}}
\newcommand{\calB}{\mbox{${\cal B}$}}
\newcommand{\calC}{\mbox{${\cal C}$}}
\newcommand{\calD}{\mbox{${\cal D}$}}
\newcommand{\calE}{\mbox{${\cal E}$}}
\newcommand{\calF}{\mbox{${\cal F}$}}
\newcommand{\calG}{\mbox{${\cal G}$}}
\newcommand{\calH}{\mbox{${\cal H}$}}
\newcommand{\calI}{\mbox{${\cal I}$}}
\newcommand{\calJ}{\mbox{${\cal J}$}}
\newcommand{\calK}{\mbox{${\cal K}$}}
\newcommand{\calL}{\mbox{${\cal L}$}}
\newcommand{\calM}{\mbox{${\cal M}$}}
\newcommand{\calN}{\mbox{${\cal N}$}}
\newcommand{\calO}{\mbox{${\cal O}$}}
\newcommand{\calP}{\mbox{${\cal P}$}}
\newcommand{\calQ}{\mbox{${\cal Q}$}}
\newcommand{\calR}{\mbox{${\cal R}$}}
\newcommand{\calS}{\mbox{${\cal S}$}}
\newcommand{\calT}{\mbox{${\cal T}$}}
\newcommand{\calU}{\mbox{${\cal U}$}}
\newcommand{\calV}{\mbox{${\cal V}$}}
\newcommand{\calW}{\mbox{${\cal W}$}}
\newcommand{\calX}{\mbox{${\cal X}$}}
\newcommand{\calY}{\mbox{${\cal Y}$}}
\newcommand{\calZ}{\mbox{${\cal Z}$}}

\newcommand*\concat{\mathbin{\|}}

\newcommand{\che}[1]{\textcolor{brown}{#1}}

\newcommand{\se}{{\it SE}}
\newcommand{\eg}{{\it e.g.}}
\newcommand{\ie}{{\it i.e.}}
\newcommand{\etal}{{\it et al.}}
\newcommand{\etc}{{\it etc}}

\newcommand{\argmin}{\mathop{\mathrm{argmin}}\limits}
\newcommand{\argmax}{\mathop{\mathrm{argmax}}\limits}

\definecolor{lightblue}{RGB}{224,236,247}
\definecolor{deepblue}{RGB}{9,46,107}
\begin{abstract}

In this paper, we propose to leverage the unique characteristics of dialogues sharing commonsense knowledge across participants, to resolve the difficulties in summarizing them. We present \textbf{SICK}, a framework that uses commonsense inferences as additional context. Compared to previous work that solely relies on the input dialogue, SICK uses an external knowledge model to generate a rich set of commonsense inferences and selects the most probable one with a similarity-based selection method. Built upon SICK, \textbf{SICK++} utilizes commonsense as supervision, where the task of generating commonsense inferences is added upon summarizing the dialogue in a multi-task learning setting. Experimental results show that with injected commonsense knowledge, our framework generates more informative and consistent summaries than existing methods. 
\end{abstract}
\section{Introduction} \label{sec:intro}

Abstractive dialogue summarization is a task of generating a shorter summary while preserving the context of a conversation~\citep{li2017dailydialog,gliwa2019samsum}. Unlike conventional document-to-document summarization (\eg, news articles and scientific publications)~\citep{nallapati2016abstractive,gehrmann2018bottom}, such \emph{dialogue-to-document} summarization suffers from the discrepancy between input and output forms, which makes learning their mapping patterns more challenging.

\begin{figure}[t!]
\centering
    \includegraphics[width=0.98\linewidth]{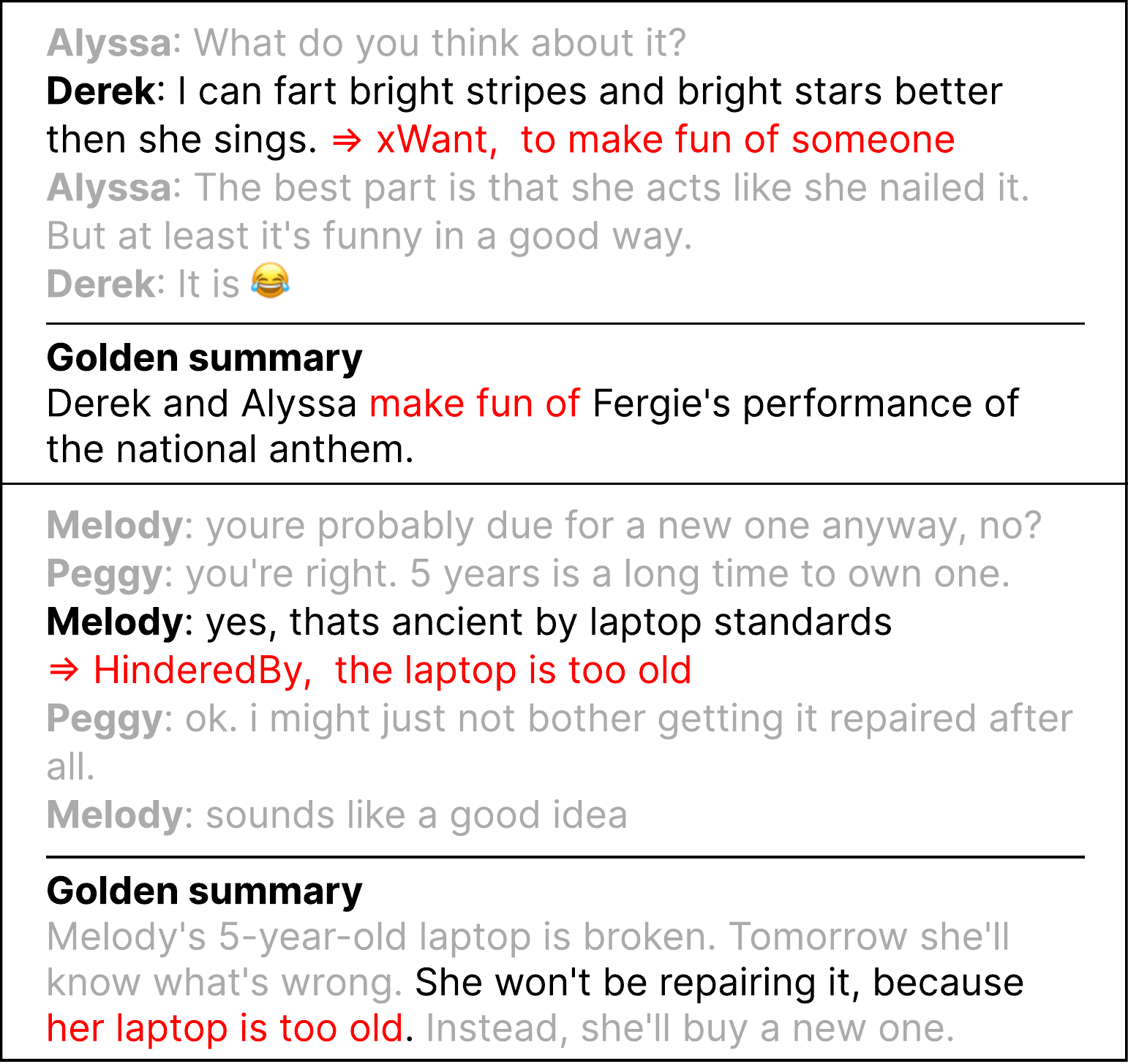}
    \caption{Example of dialogue-summary pairs. Capturing the intention and hidden meaning is important to generate a novel summary.}
\label{fig:motivation}
\end{figure}

There are two key challenges that make summarizing dialogues harder than documents. First, detecting unspoken intention is crucial for understanding an utterance~\citep{mendelsohn1994learning, ram2018conversational}. As shown in Figure~\ref{fig:motivation}, without understanding the intent ``\emph{to make fun of someone}'', it is hard to write a correct summary. Second, there exists information that can only be understood when its hidden meaning is revealed~\citep{talmy1988force}. For example, it is important to capture the hidden meaning ``\emph{The laptop is too old}'' beyond the written text ``\emph{yes, thats ancient by laptop standards}'' when writing the summary.
%Second, since people do not state the obvious~\citep{grice1975logic}, it is challenging to understanding the overall conversational flow with sparse amount of information.

% understanding the intent ''\emph{to make fun of someone}'' is critical when generating a summary.
%As shown in Figure 1, we observe that the lexical overlap between input and output tokens is consistently lower in the dialogue datasets (blue) than in the general document datasets (red). It is because informative contents needed for a desirable summary are not explicitly stated in dialogues (Figure 1b) or, even if explicitly stated, their colloquial form hinders aligning with the relevant part of the gold summary (Figure 1c). 

\begin{comment}
\begin{figure}[t!]
     \centering
     \begin{subfigure}[t]{0.44\textwidth}
         \includegraphics[width=\textwidth]{image/overlap_figure.pdf}
         \caption{Averaged number of overlapped tokens}
         %\label{fig:scene1}
     \end{subfigure}
     \vfill
     \begin{subfigure}[t]{0.44\textwidth}
         \includegraphics[width=\textwidth]{image/motivation.pdf}
         \caption{Example dialogues that require commonsense to be understood}
         %\label{fig:scene2}
     \end{subfigure}
\caption{A set of illustrative examples that represents a role of commonsense knowledge.}
\label{fig:intro}
\end{figure}
\end{comment}

\begin{comment}
\begin{figure}[t!]
\centering
    \includegraphics[width=1\linewidth]{image/overlap_figure.pdf}
    %\includepdf[]{image/motivation.pdf}
    \caption{Averaged number of overlapped tokens observed on dialogue summarization datasets and document summarization datasets.}
\label{fig:motivation}
\end{figure}
\end{comment}

\begin{comment}
\begin{figure}[t!]
    \centering
    % \includegraphics[width=1\linewidth]{image/figure1_2x.png}
    \includepdf[pages=-]{image/figue1.pdf}
    \caption{Examples of dialogue summary generated by BART and SICK: BART fails to capture hidden contents from dialogue but SICK is capable of generating abundant summary with commonsense knowledge.}
    \label{fig:intro}
\end{figure}
\end{comment}
Commonsense knowledge models~\citep{hwang2021comet,gabriel2021paragraph,west2021symbolic} such as COMET can generate a set of event-centered (\eg, \textsc{HinderedBy}, \textsc{xReason}, \textsc{xNeed}) and social-interaction (\eg, \textsc{xIntent}, \textsc{xWant}) commonsense inferences. 
We argue that the aforementioned issues can be mitigated using commonsense knowledge by \textit{filling in the gap} in a dialogue.
% event-centered, social interaction commonsense knowledge 가 missing gap을 채워줄 수 있다.

Despite its effectiveness, it is non-trivial to use commonsense knowledge for improving abstractive dialogue summarization performance. While commonsense knowledge has been widely applied to commonsense reasoning~\citep{bosselut2019dynamic,liu2020commonsense,chang2021incorporating,wang2021retrieval,kimconvei} or question answering~\citep{shwartz2020unsupervised,bosselut2021dynamic}, its usage for summarization is understudied~\citep{feng2021incorporating}.

In this paper, we present our framework \textbf{SICK} and its extension \textbf{SICK++} to properly inject commonsense knowledge into state-of-the-art language models (\eg, BART~\citep{lewis2019bart}) for abstractive dialogue summarization. We argue a na\"ive adoption of commonsense only hurts performance in summarization, as (a) expanding source contents is counter-intuitive approach for the goal of condensation, and (b) simply adding additional inputs in pre-trained language models does not lead to robust inferences as reported in~\citet{zhou2020rica,zhou2021probing}. Our framework addresses this by (a) filtering and (b) robust training. 

Based on analytical measurements, commonsense knowledge is selected and enumerated as an additional context of dialogue inputs. In SICK++, we also design a new auxiliary task named \emph{commonsense supervision}. Using commonsense knowledge generated from gold summaries as additional supervision, the goal of the task is to generate the target commonsense. Then, the dialogue summarization and commonsense generation tasks are jointly learned in a multi-task learning setting to effectively inject commonsense knowledge into the shared encoder.

%That is, such multi-task learning aims to enforce the model to store and aggregate the different types of dialogue and commonsense knowledge into the same embedding space.

\begin{figure*}[t!]
\centering
    \includegraphics[width=0.95\linewidth]{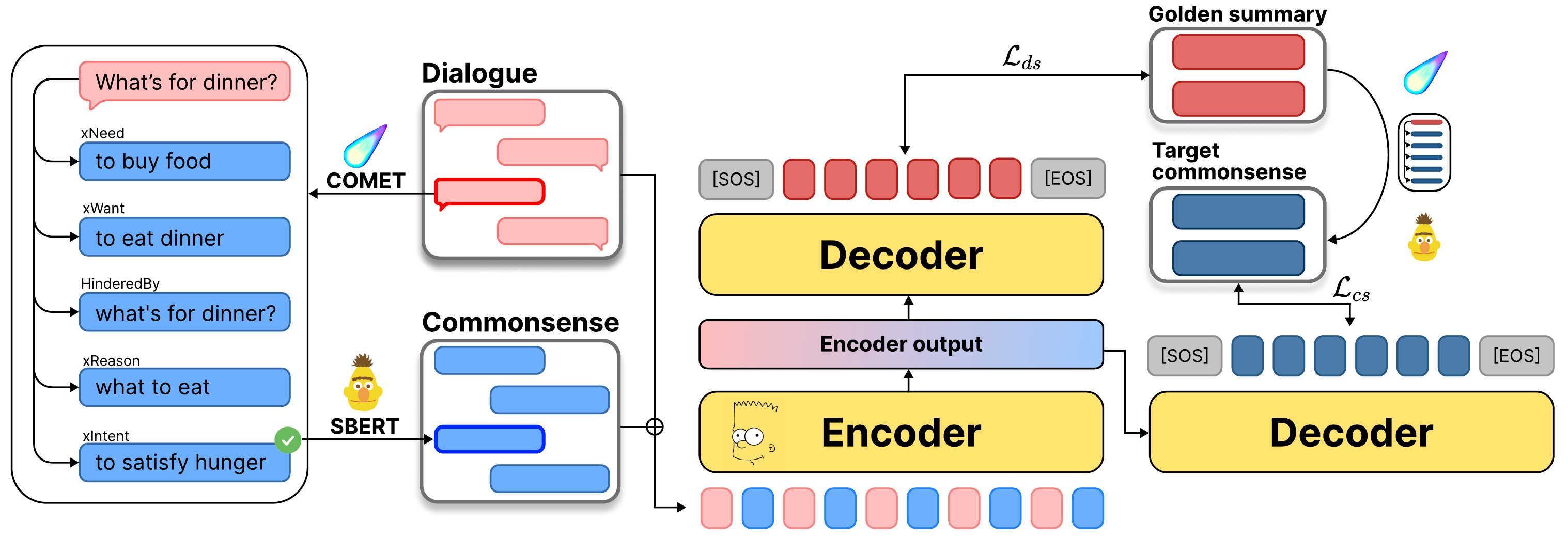}
    \caption{The overall framework of SICK and SICK++. The decoder generating target commonsense is used for SICK++.}
    \label{fig:overview}
\end{figure*}

% To Do
To validate our framework, we conduct a set of experiments on abstractive dialogue summarization benchmarks. Empirical results show that our framework can improve summarization performance with leveraged commonsense knowledge, outperforming other baselines. Human evaluation results prove that our method can generate informative and consistent summaries. In addition, we conduct experiments to analyze the effect of commonsense knowledge on abstractive dialogue summarization.
 
%In addition, our method shows good performance in zero-shot and low-resource settings. Lastly, we also analyze different commonsense knowledge and its effects on abstractive dialogue summarization.

%Lastly, we show that our method is also effective in zero-shot or low resource settings.
%%Moreover, we find that the output from our model obtains quality improvement over baselines based on human evaluation. 

\section{Related Work}

\subsection{Abstractive Dialogue Summarization}

Compared to extractive summarization~\citep{nallapati2017summarunner,zhang2018neural,zhong2020extractive}, abstractive summarization is considered more challenging and has received extensive attention~\citep{rush2015neural,see2017get}. Benefiting from the advance of large-scale pre-trained language models, the performance of encoder-decoder models has achieved substantial improvements in document summarization~\citep{nallapati2016abstractive,gehrmann2018bottom,zhang2020pegasus}.
% generally, extensive

Recently, abstractive dialogue summarization has 
become another emerging research area, where the goal is to generate concise summaries for conversations such as meetings~\cite{zhu2021mediasum} and chit-chat~\cite{chen2021dialogsum}.
% Compared to formal documents, dialogues are more difficult to be 
% Compared to formal documents, dialogues tend to be more abstract, because people do not necessarily state the obvious~\cite{grice1975logic}. Furthermore, unlike formal documents, conversations have a more dynamic and interactive flow of information exchange between speakers~\cite{li2021conversations}, which are often informal, verbose, and repetitive.
It is more difficult to capture the key points in dialogues than documents, because people do not state the obvious~\citep{grice1975logic} and conversations have a more interactive flow of information between speakers~\cite{li2021conversations}.
Based on the characteristic of the dialogues, many studies focused on organizing the information in the dialogues. \citet{wu2021controllable} propose to create a summary sketch for a given dialogue as weak supervision. \citet{chen2021structure} explicitly model structures in conversations by incorporating discourse relations and action triples in utterances through structured graphs. %\citet{feng2021incorporating} propose to leverage commonsense for building heterogeneous dialogue graph to get better understanding of a structure of the dialogue. 
Instead of organizing the given dialogue for better understanding, our method adds additional knowledge to fill in the missing cues between dialogues.
% A vew studies have also paid attention to utilizing conversational analysis for generating dialogue ssummaries, such as 

%while \citet{feng2021language} uses context-response pairs and keywords as annotation.
%\citet{lei2021finer} proposes to utilize dialogue semantic structures,
%While it is a chellenging task, with large-scale datasets and sophisticated neural architectures, the performance of abstractive models have achieved substantial improvements in the news domain, especially benefiting from the advance of pre-trained language models with encoder-decoder(Seq2Seq) architecture. 
%This study has a lot in common with our study in the sense that it made use of commonsense on abstract dialogue summarization tasks. However, there is a decisive difference in that they only used commonsense as additional auxiliary information. 
% TV series, interview,
%Based on the characteristics of the conversations, many studies pay attention to utilizing conversational analysis for dialogue summarization, such as leveraging dialogue acts, multi-modal features, topic information, and fine-grained view segmentation with hierarchical modeling. 
%It is critical to find out what information is needed to fully understand the conversation between people.
%In other words, an extra auxiliary task of filling in the implicit information is needed to solve abstractive dialogue summarization. 
% such as leveraging dialogue acts, multi-modal features, topic information, and fine-grained view segmentation with hierarchical modeling. 

\subsection{Commonsense Knowledge Models}\label{sec:layout}
% kg를 사용하는 방식 <-> lm을 사용하는 방식
Recent research has focused on commonsense knowledge acquisition through different lines: commonsense knowledge graphs and commonsense knowledge models. Unlike static knowledge graphs such as ATOMIC~\citep{sap2019atomic} in which entities and relations between entities are represented in nodes and edges, commonsense knowledge models such as COMET~\cite{bosselut2019comet} have been shown to generate implicit commonsense inferences along several dimensions depending on what knowledge graphs they are trained on. Commonsense knowledge models can be used to anticipate and reason unobserved causes and effects in relation to the observed event~\cite{sap2019atomic}. Despite these functions, they are applied on defined domains~\cite{shwartz2020unsupervised,bosselut2021dynamic}. Especially, on dialogue summarization task, there has been limited usage of using commonsense directly as additional context. For example, \citet{feng2021incorporating} and \citet{zhou2022think} utilized ConceptNet~\cite{speer2017conceptnet}, a static knowledge graph with encyclopedic knowledge, to fill in the missing cues between dialogue. 

In contrast to encyclopedic knowledge, our method uses event-centered and social-interaction knowledge as additional context. Also, instead of retrieving from a static knowledge graph, our method deploys on-the-fly commonsense knowledge models to acquire a rich set of commonsense inferences dynamically.

\section{Proposed Framework}\label{sec:sick}

% We now present the SICK framework that generates and leverages commonsense knowledge to advance abstractive dialogue summarization. Suppose that a dialogue $\calD = \{ u_1, u_2, ..., u_n \}$ is given with $n$ utterances, coupled with its corresponding summary $\calY = \{ y_1, y_2, ..., y_m \}$ where $m$ is the number of sentences. We frame such task as a sequence-to-sequence learning problem, and thus the goal is to learn a model $\mathbb{M} : \calD \rightarrow \calY$ with encoder and decoder, which generates a novel summary from a dialogue. Figure~\ref{fig:overview} illustrates the overall architecture of the SICK framework.

%Our ultimate goal is to advance abstractive dialogue summarization by leveraging commonsense knowledge, as llustrated in Figure~\ref{fig:overview}. 
In this section, to inject commonsense knowledge for rich abstractive dialogue summarization, we introduce our new framework, SICK(\underline{S}ummarizing with \underline{I}njected \underline{C}ommonsense \underline{K}nowledge) and its extension SICK++, as shown in Figure~\ref{fig:overview}. 

%We define the task formulation(Section \ref{task description}), explain how to obtain a set of commonsense knowledge (Section \ref{commonsense knowledge generation}) and describe our dual-decoder model architecture (Section \ref{sec:sick.1}, \ref{sec:sick.2}).

\subsection{Task Description}
Our task definition follows a sequence-to-sequence learning problem setting. Based on pre-trained generative language models, our goal is to learn a mapping function $\mathbb{M} : \calD \rightarrow \calY$ where $\calD = \{ u_1, u_2, ..., u_n \}$ is a dialogue with $n$ utterances, and $\calY = \{ y_1, y_2, ..., y_m \}$ is a corresponding summary of $m$ sentences. 

We further extend the task with two modifications. First, we generate and filter to acquire a set of commonsense knowledge $\calC = \{ c_1, c_2, ..., c_n \}$ based on $\calD$ (Section \ref{commonsense knowledge generation}, \ref{sec:sick.1}). 
Then, we adjust the mapping function as $\mathbb{M} : \calX \rightarrow \calY$, where $\calX$ is a cross concatenation of $\calD$ and $\calC$ (Section \ref{sec:sick.1}). Second, we add an auxiliary task \textit{commonsense supervision}, $\mathbb{M}^* : \calX \rightarrow \calZ$, where the target commonsense $\calZ = \{ z_1, z_2, ..., z_m \}$ is acquired based on $\calY$ (Section \ref{sec:sick.2}).

%We adopt an encoder-decoder model, BART~\cite{lewis2019bart} as our base architecture, where the encoder and decoder consists of a stack of Transformer layers respectively.

%Suppose that a dialogue $\calD = \{ u_1, u_2, ..., u_n \}$ is given with $n$ utterances, coupled with its corresponding summary $\calY = \{ y_1, y_2, ..., y_m \}$ where $m$ is the number of sentences. We frame such task as a sequence-to-sequence learning problem, and thus the goal is to learn a model $\mathbb{M} : \calD \rightarrow \calY$ with encoder $\mathbb{E}$ and decoder $\mathbb{D}$, which generates a novel summary from a dialogue. We adopt BART~\cite{lewis2019bart} as our base architecture, where encoder and decoder consists of a stack of Transformer layers respectively.
%SICK is a framework that leverages commonsense knowledge to advance abstractive dialogue summarization, as illustrated in Figure~\ref{fig:overview}. 

%\subsection{Task Formulation}

% Our ultimate goal is to enhance abstractive summaries of dialogues by leveraging commonsense knowledge into dialogue summarization.

%% To ground each dialogue to commonsense knowledge, we make use of conceptnet to incorporate knowlege.
%% make full use of large-scale commonsense knowledge
%%% We propose a model named xx for incorporating commonsense knowledge by constructing the graph including both utterance and knowledge nodes.

%Due to these characteristics, a model is required to reason about context and infer information not explicitly expressed, which makes abstractive dialogue summarization more challenging~\citep{khalifa2021abagof}.

\subsection{Commonsense Knowledge Generation}
\label{commonsense knowledge generation}

% 1문단 : 정보 언급 없음, 있더라도 폼이 달라서 매핑이 어려움
% 2문단 : 위 문제를 해결하기 위해 commonsense를 summarization에 활용하자! for example, 커먼센스가 도움 되는 예시 두개 / 
% 3문단 : For that, external model 사용하기로 함, input output 정의 / Specifically, COMET and PARACOMET / relation type 많고 각 예시 들기

%In conversations, people tend to optimize against stating the obvious~\citep{grice1975logic}. We hypothesize that it is non-trivial for a model to fully understand the dialogue with only utterances on the surface. 
% We propose to leverage commonsense knowledge as a supplement to insufficient dialogue contexts.
In SICK, commonsense knowledge is leveraged as a supplement to insufficient context of dialogues. As shown in Table~\ref{tab:relation_comet}, additional information can be derived from the given utterance in various aspects. There are some cases where the intention of the speaker is crucial in comprehending the dialogue (\eg, ``\emph{to believe in something}'', ``\emph{to talk to someone about dreams}''). Whereas in other cases, the hidden information is
necessary (\eg, ``\emph{Charlie doesn't believe in dreams}'', ``\emph{to have a dreams}'', ``\emph{Charlie is a skeptic}'').  We adopt an external commonsense knowledge model that generates a diverse and abundant set of commonsense inferences in natural language. Given a text $x$ and a relation type $r$, the commonsense knowledge model gives an output $c$ grounded to the relation type. \ie, $f: (x, r) \rightarrow c$. 
%For example, a commonsense ``\emph{Julia looks at his son}''  that is generated by an utterance ``\emph{Is that your son?}'' fills in the missing information, \ie, speakers share a photo of Hugh's son, which is not explicitly mentioned in the flow of conversation, when generating the summary ``\emph{Hugh shares a photo of his son with Joan and Julia.}''. As an another example, an utterance ``\emph{I can't wait to vote for anyone else but him}'' with informal language is able to be transformed to a commonsense ``\emph{They vote for someone else}'' with formal language, which is helpful for generating summary ``\emph{They hope he gets voted out.}''.

\begin{table}[t!]
\begin{center}\small
{\begin{tabular}{l l}
    \toprule
    \textbf{Utterance} & \underline{Charlie} : Do you really believe \\
    & that dreams can mean something? \\
    \midrule
    \textsc{HinderedBy} & Charlie doesn't believe in dreams.\\
    \textsc{xWant} & to talk to someone about dreams. \\
    \textsc{xIntent} & to believe in something. \\
    \textsc{xNeed} & to have a dream. \\
    \textsc{xReason} & Charlie is a skeptic. \\
    \bottomrule
\end{tabular}}
\end{center}
\caption{Example of commonsense knowledge generated by COMET given a dialogue.}
\label{tab:relation_comet}
\end{table}
% \begin{table}[t!]
% \begin{center}\small
% {\begin{tabular}{l l}
%     \toprule
%     \textbf{Utterance} & \underline{Sophie} : Maybe Alex put it somewhere. \\
%     & that dreams can mean something? \\
%     \midrule
%     \textsc{HinderedBy} & They don't know where it is.\\
%     \textsc{xWant} & to find it. \\
%     \textsc{xIntent} & to believe in something. \\
%     \textsc{xNeed} & to have a dream. \\
%     \textsc{xReason} & they are looking for something. \\
%     \bottomrule
% \end{tabular}}
% \end{center}
% \caption{Example of commonsense knowledge generated by COMET given a dialogue.}
% \label{tab:relation_comet}
% \end{table}

%To this end, we adopt an external commonsense knowledge model which generates diverse and abundant commonsense descriptions in natural language. Given a text $x$ and a relation type $r$, the commonsense knowledge model outputs a commonsense inference $c$ grounded to the relation type, \ie, $f: (x, r) \rightarrow c$. 

Specifically, we use COMET~\citep{hwang2021comet}, a widely-used generative commonsense model as our external model. Among the 23 possible candidate relation types, we choose 5 unique relations that helps understand the speakers' intentions and find out the missing information. COMET generates 5 commonsense inferences per relation type, resulting in 25 per input. %For example, given an utterance ``\emph{The toilet paper is finished, could you fetch me some tissues, please?}'', COMET generates ``\emph{PersonX doesn't have any tissues.}'' for relation type \textsc{HinderedBy}, while ``\emph{PersonX wants to use the toilet}'' for relation type \textsc{xWant}, capturing different aspects of the utterance for better understanding the dialogue.

% However, generating commonsense inferences only relying on each utterance might cause another problem of not considering episodic knowledge.
Also, to attend to the previous utterances when generating commonsense inferences, we further explore a discourse-aware model, PARA-COMET~\citep{gabriel2021paragraph} that generates coherent inferences. More specifically, while COMET generates a set of commonsense inferences considering only one sentence at a time, PARA-COMET adopts an internal memory module to consider previous dialogue history when generating an output. 
%Given a list of utterances and relation types, PARA-COMET outputs a commonsense candidate for each sentence, which is consistent with the entire narrative flow of the input text.

\begin{table}[t!]
\begin{center}\small
{\begin{tabular}{l l}
    \toprule
    \multirow{2}{0.1em}{\textbf{Prev-Utterances}} & \underline{Jane} : google maps says it is at least 3h \\
    & \underline{Steven} : I used to make it in 2, trust me \\
    & \underline{Jane} : but it's almost 300km\\
    & \underline{Steven} : the road is new , we will make it\\
 
    \textbf{Utterance} & \underline{Jane} : I don't want to stress out, let's \\
    & \;\;\;\;\;\;\;\;\;\;meet at 4:30 instead of 5, ok? \\
    \midrule
    \textsc{xIntent} & to avoid stress. \\
    \textsc{xWant} & \textbf{to not be late.} \\
    \textsc{xReact} & annoyed \\
    \textsc{xEffect} & PersonX sweats from nervousness. \\
    \textsc{xAttr} & nervous. \\
    \bottomrule
\end{tabular}}
\end{center}
\caption{Example of commonsense knowledge generated by PARA-COMET given a dialogue.}
\label{tab:relation_paracomet}
\end{table}
% Vesna: Well doneeeee!.
% Ost: There is a big crowd in parking, so the prices for finding garage places are favorable, in this area.
% Vesna: Really!.
% Ost: Return on investment is much higher than if I give money under term savings in the Bank.
% Vesna: You make sense for money.
% Jane: google maps says it is at least 3h <file_other>
% Steven: I used to make it in 2, trust me :D
% Jane: but it's almost 300km..
% Steven: the road is new , we will make it ^^
% Jane: I don't want  to stress out , let's meet at 4:30 instead of 5, ok?

In Table~\ref{tab:relation_paracomet}, when generating commonsense inferences of the current dialogue, PARA-COMET conditions on the previous utterance. Knowing what was previously stated, the intention of the speaker (\eg, ``to not be bothered'', ``to not be stressed'', ``upset'') and the hidden knowledge (\eg, ``annoyed'', ``PersonX gets into trouble'') differs from COMET.

\subsection{Summarizing with Injecting Commonsense Knowledge (SICK)}\label{sec:sick.1}

% input으로 써야 하는 이유 : 공유된
% target으로 써야 하는 이유

% 1문단 : commonsense를 input으로 쓰기로 함 but --> 모든 commonsense를 쓰면 길이 늘어나서 summarization에 악영향 있음 -->
% 2문단 : sematic적으로 가장 비슷한 걸 고르자

\vspace{0.8mm}
\noindent\textbf{Filtering} Compared to question answering and commonsense generation~\citep{shwartz2020unsupervised,wang2021retrieval}, summarizing dialogues has another difficulty. The amount of data provided as the input should be mapped into the output in a concise form. Therefore, simply providing extra input (\ie, commonsense knowledge) may confuse the model when generating a summary. Moreover, it is unable to add every possible commonsense knowledge to the dialogue due to the limited input sequence length of transformer-based models.%Therefore, it is trivial to set a standard in order to filter out and select potentially important commonsense as opposed to simply concatenate every possible choices. 

To address this issue, we propose to select the most favorable commonsense for each utterance. For 25 candidates, we measure the semantic relevance between the utterance and the commonsense inference concerning. One could imagine that filtering could choose only very similar ``safe'' examples that might not be as valuable/interesting in practice (\ie, \textit{diversity vs. quality}). However, recent literature address that paradoxically, filtering increases diversity~\citep{west2021symbolic}. We also discuss the impact of different filtering methods in Appendix~\ref{sec:roberta}.

We employ SBERT~\citep{reimers2019sentencebert} to compute the similarity score between utterance and commonsense pairs. We select one commonsense inference $c_{i}$, with the highest score for each utterance $u_i$ among the candidate relations $\calR$. As a result, we obtain the input commonsense ${\calC} = \{ c_i \}_{i=1}^n$ aligned with dialogue $\calD$.
\begin{equation}
    c_i = \argmax_{ {c_{i}^{r}}}(\text{score}(u_i, {c_{i}^{r}}))\quad 
    (r \in \mathcal{R}) %\quad l \in [1, L]
\end{equation}

\vspace{0.8mm}
\noindent\textbf{Cross Concatenation} After obtaining the input commonsense for the dialogue, we concatenate the dialogue and its corresponding set of commonsense inferences. To encode the information that $c_i$ is derived from $u_i$, we enforce to attend its neighbor token. Instead of concatenating $\calD$ and $\calC$ back and forth, we concatenate turn by turn considering \textit{locality of reference}~\citep{clark2019does,zaheer2020big}, where tokens tend to attend its neighboring tokens. In order to separate the modalities between dialogues and commonsense inferences, we add special tokens \textsc{<I>}, \textsc{</I>} in back and forth of each commonsense inference $c_i$. Thus the input sequence $\calX$ is formulated as:
\begin{equation}\label{eq:input_format}
    \calX = \calD \oplus {\calC} = \cdots \parallel u_i\parallel \textsc{<I>}\, c_i\, \textsc{</I>} \parallel \cdots
\end{equation}

\vspace{0.8mm}
\noindent\textbf{Training} SICK is built upon a transformer-based encoder-decoder architecture. The encoder fuses the information from two different modalities (\ie, dialogue and commonsense inference). By the stack of decoders, the encoder output is used for cross-attention with the summary. The training objective, a negative-log likelihood parameterized by $\theta_{ds}$, can be formulated as:
\begin{equation}\label{ds_loss}
    \calL_{ds} = - \sum_{i=1}^{|\mathcal{Y}|} \sum_{j=1}^{|y_i|} \log P(w_{i,j}|w_{i<j}, \calX; \theta_{ds}) 
\end{equation}
where $w_{i, j}$ is $j$-th token of $i$-th sentence $y_i$ in target summary $\calY$.

\subsection{SICK++}\label{sec:sick.2}

%Based on the output of the last layer of $\mathbb{D}_{ds}$, we can get the generated summary $\hat{\calY}$, whereas we can get the predicted commonsense $\hat{\tilde{\calY}}$ based on the output of the last layer of $\mathbb{D}_{cr}$. %The loss for dialogue summarization decoder $\mathbb{D}_{ds}$ is defined as the negative log-likelihood:

%To adopt commonsense knowledge as additional supervision, we formulate an auxiliary task of commonsense generation, and thus employ dual-decoders, \ie, dialogue summarization decoder $\mathbb{D}_{ds}$ and commonsense generation decoder $\mathbb{D}_{cg}$, while maintaining the shared encoder. 

%To provide supplement for insufficient dialogue context, we propose to incorporate commonsense knowledge $\tilde{\calD}$ at input side, which is obtained from a learned commonsense knowledge model given a dialogue $\calD$. By using additional input commonsense, the model benefits from the explicit contents beyond dialogue as well as dialogue context itself.
%employ dual-decoders, not only dialogue summarization decoder $\mathbb{D}_{ds}$ but also commonsense generation decoder $\mathbb{D}_{cg}$ to make full use of commonsense knowledge

%where $c_r^{y_i}$ is a commonsense for the sentence $y_i$ in summary $\calY$, and $r$ is a relation type. 

% 지식이 있는 것과 지식을 활용하는 것 은 다르닷
% 모델에게 input으로 cs를 그냥 던져준다고 해서 이를 summary에 활용한다는 보장이 없다.
% 상식을 inject한다고 해서 무조건 summary task에 도움이 되지 않는다. 즉 상식이 inject되고 이가 summary 에 도움이 되는 auxiliary task라는 점을 어필해야함
% dialog에서 생성할 수 있는 cs와 summary에서 생성할 수 있는 cs의 차이점은 무엇일까?
% => summary에서 생성할 수 있는 cs는 대화의 맥락을 이해해야 생각할 수 있는 cs가 존재함
\noindent\textbf{Commonsense Supervision} It is well known that models do not consider the actual input as a whole and only look at certain parts of the input therefore not performing the underlying task but some derivative~\citep{branco2021shortcutted}. For example, in Figure~\ref{fig:motivation}, although it is critical to understand Derek's intention (\eg, ``\emph{to make fun of Fergie's performance}''), SICK may not utilize the commonsense to comprehend the dialogue. 

To overcome this problem, we propose an auxiliary task named \textit{commonsense supervision}. In addition to providing commonsense on the input side, we also leverage commonsense knowledge as additional target variable, which prevents the model from disregarding commonsense and enforces actually to utilize commonsense. For instance, when the summary ``\emph{Derek and Alyssa make fun of Fergie's performance of the national anthem.}'' is given to COMET, we observe that a target commonsense ``\emph{to make fun of}'' is generated. Generating both the summary and the target commonsense has an effect of emphasizing that the input commonsense inference ``\emph{to make fun of someone}'' is important.

%  Furthermore, in abstractive dialogue summarization task, it is non-trivial for model to incorporate different types of information, \ie, informal dialogues and formal commonsense inferences, for whole input at once~\citep{khalifa2021abagof}.

We generate a set of target commonsense inferences $\calZ$ with the summary $\calY$ using an external knowledge model $f$. Then we filter and select the most plausible target commonsense.
\begin{equation}
    z_i = \argmax_{{z_{i}^{r}}}(\text{score}(y_i, {z_{i}^{r}}))\quad 
    (r \in \mathcal{R}) %\quad l \in [1, L]
\end{equation}

To adopt commonsense knowledge as additional supervision, we further include commonsense summarization decoder $\mathbb{D}_{cs}$, which learns to generate target commonsense $\calZ$.

\vspace{0.8mm}
\noindent\textbf{Training} With the target commonsense $\calZ$, we train the commonsense summarization decoder $\mathbb{D}_{cs}$ to minimize a negative log-likelihood loss function such as:
\begin{equation}\label{cs_loss}
    \calL_{cs} = - \sum_{i=1}^{|\calZ|} \sum_{j=1}^{|z_i|} \log P(w_{i,j}|w_{i<j}, \calX ; \theta_{cs}) 
\end{equation}
where $w^i_j$ is a $j$-th word token of sentence $c^y_i$ from the target commonsense $\calZ$. 

We linearly combine the two loss functions, Equation~\ref{ds_loss} and Equation~\ref{cs_loss}, in a multi-task learning setting as follows:
\begin{equation}
    \calL_{total} = \lambda \cdot \calL_{ds} + (1 - \lambda) \cdot \calL_{cs}
\end{equation}
where $\calL_{ds}$ and $\calL_{cs}$ denote the loss function for dialogue summarization decoder $\mathbb{D}_{ds}$ and commonsense summarization decoder $\mathbb{D}_{cs}$, respectively. $\lambda$ is a predefined hyperparameter to adjust the scale of each loss. In our setting, we set $\lambda = 0.66$.
% 마찬가지로 theta_cs에 대한 설명

\vspace{0.8mm}
\noindent\textbf{Inference} During inference, given an input dialogue $\calD_{test}$, we first obtain input commonsense $\calC_{test}$ for the dialogue, and specify input sequence as $\calX_{test} = \calD_{test} \oplus \calC_{test}$ by concatenating turn by turn. Then, the model predicts summary $\hat{\calY}_{test} = \mathbb{M}(\calX_{test})$ for the dialogue $\calD_{test}$. Note that while we train the model in a  dual-decoder setting, we only use the dialogue summarization decoder $\mathbb{D}_{ds}$ and discard the commonsense prediction decoder $\mathbb{D}_{cs}$ at inference time.

\section{Experimental Setup}

\begin{table}[t!]
\begin{center}\small
{\begin{tabular}{l c c}
    \toprule
    & SAMSum & DialogSum \\
    \midrule
    Train & 14,732 & 12,460 \\
    Dev & 818 & 500 \\
    Test & 819 & 500 \\
    \#Tokens/dialogue & 82.57 & 121.56 \\
    \#Tokens/summary & 20.30 & 22.64 \\
    \#Turns & 11.2 & 9.5 \\
    \#Speaker & 2.4 & 2.0 \\
    \#Compression rate & 0.3538 & 0.2001 \\
    \bottomrule
\end{tabular}}
\end{center}
\caption{Statistics of dialogue summarization datasets. \# stands for the average number. The compression rate is a ratio of the length of summary divided by the length of dialogue.}
\label{tab:dataset_stat}
\end{table}

\begin{table*}[t!]
\begin{center}\small
{\begin{tabular}{l cccc cccc}
    \toprule
    & \multicolumn{4}{c}{SAMSum} & \multicolumn{4}{c}{DialogSum} \\
    \cmidrule(lr){2-5} \cmidrule(lr){6-9}
    Model & R-1 & R-2 & R-L & B-S & R-1 & R-2 & R-L & B-S  \\
    \midrule
    PointerGenerator~\citep{see2017get}$^{\ast}$ & 32.27 & 14.42 & 34.36 & / & / & / & / & /\\
    DynamicConv~\citep{wu2019pay}$^{\ast}$ & 41.07 & 17.11 & 37.27 & / & / & / & / & /\\
    Transformer~\citep{vaswani2017attention}$^{\ast}$ & 42.37 & 18.44 & 39.27 & / & / & / & / & /\\
    DialoGPT~\citep{zhang2019dialogpt}$^{\dagger}$ & 39.77 & 16.58 & 38.42 & / & / & / & / & /\\
    BART-xsum \cite{lewis2019bart}$^{\dagger}$ & 51.74 & 26.46 & 48.72 & 53.87 & / & / & / & /\\
    UniLM~\citep{dong2019unified}$^{\dagger}$ & 47.85 & 24.23 & 46.67 & / & 42.38 & 16.88 & 34.36 & 69.40\\
    PEGASUS~\citep{zhang2020pegasus}$^{\dagger}$ & 50.50 & 27.23 & 49.32 & 53.35 & 38.40 & 13.84 & 33.41 & 68.20\\
    BART-xsum~\citep{lewis2019bart}$^{\ddag}$ & 52.50 & 27.67 & 48.75 & 68.16 & 45.15 & 19.78 & 36.57 & 71.09\\
    % bart-xsum bert score : 68.16(ours; reimplemented)
    \midrule
    D-HGN~\citep{feng2021incorporating} & 42.03 & 18.07 & 39.57 & 64.20 & / & / & / & /\\
    S-BART~\citep{chen2021structure} & 50.70 & 25.50 & 48.08 & 70.07 & / & / & / & / \\
    CODS~\citep{wu2021controllable} & 52.65 & 27.84 & \textbf{50.79} & 66.55 & 44.27 & 17.90 & 36.98 & 70.49\\
    % new_weights_relation_samsum/final_context_Trial1_xNeed.txt (samsum), new_weights_comet_dialogsum/final_Trial1_context_comet.txt (dialogsum)
    % /home/intern2/ICSK4AS/src/new_weights_comet/final_Trial1_context_comet/xNeed/result.txt (samsum), new_weights_paracomet_dialogsum/trial/xEffect/result.txt (dialogsum) /home/intern2/ICSK4AS/src/new_weights_paracomet_dialogsum/trial/xEffect/checkpoint-4665/result.txt,
    \midrule
    SICK \quad w/ COMET (Ours) & 53.04 & 27.60 & 48.49 & 71.61 & 45.70 & 20.08 & 40.26 & 71.08 \\
    SICK++ w/ COMET (Ours) & 53.24 & 28.10 & 48.90 & 71.71 & \textbf{46.26} & \textbf{20.95} & \textbf{41.05} & 71.30 \\
    \midrule
    SICK \quad w/ PARA-COMET (Ours) & 53.39 & 28.42 & 49.12 & 71.83 & 46.01 & 20.30 & 40.75 & \textbf{71.57} \\
    SICK++ w/ PARA-COMET (Ours) & \textbf{53.73} & \textbf{28.81} & 49.50 & \textbf{71.92} & 46.20 & 20.39 & 40.83 & 71.32 \\
    \bottomrule
\end{tabular}}
\end{center}
\caption{Automatic evaluation on abstractive dialogue summarization benchmarks, \ie, SAMSum and DialogSum. Results on SAMSum with * are obtained from \cite{gliwa2019samsum}, $\dagger$ are obtained from \cite{wu2021controllable} and ${\ddag}$ is a re-implemented version trained under the same conditions with ours for fair comparison. Results on DialogSum for all models are all reimplemented under the same conditions with ours.}
%Results on DialogSum with $\dagger$ are re-implemented with \cite{wolf2020transformers} under same conditions with our models.
\label{tab:automatic_evaluation}
\end{table*}

\begin{comment}
    SICK w/ COMET & 53.18 & 27.93 & 49.05 & 0.7161 & 44.55 & 19.14 & 39.60 & ? \\
    SICK++ w/ COMET & 53.49 & 28.25 & 48.95 & 0.7167 & \textbf{47.46} & 21.76 & 42.22 & 0.7155 \\
    % Google Drive (samsum), new_weights_paracomet_dialogsum/trial/xEffect/result.txt (dialogsum)
    SICK w/ PARACOMET & 53.42 & 28.20 & 49.06 & 0.7188 & 45.78 & 20.27 & 40.43 & ? \\
    SICK++ w/ PARA-COMET & \textbf{53.53} & \textbf{28.69} & 49.23 & \textbf{0.7194} & 47.32 & \textbf{21.97} & \textbf{42.45} & \textbf{0.7158} \\
    
    %%%%% FINAL PERFORMANCE %%%%%%
    SICK \quad w/ COMET (Ours) & 53.04 & 27.60 & 48.49 & 71.61 & 45.70 & 20.08 & 40.26 & 71.08 \\
    SICK++ w/ COMET (Ours) & 53.24 & 28.10 & 48.90 & 71.71 & \textbf{46.26} & \textbf{20.95} & \textbf{41.05} & 71.30 \\
    SICK \quad w/ PARACOMET (Ours) & 53.39 & 28.42 & 49.12 & 71.83 & 46.01 & 20.30 & 40.75 & \textbf{71.57} \\
    SICK++ w/ PARACOMET (Ours) & \textbf{53.73} & \textbf{28.81} & 49.50 & \textbf{71.92} & 46.20 & 20.39 & 40.83 & 71.32 \\
\end{comment}

\subsection{Datasets and Baselines}

We perform experiments on SAMSum~\citep{gliwa2019samsum} and DialogSum~\citep{chen2021dialogsum} datasets. SAMSum is the most widely used resource for abstractive dialogue summarization task. It consists of natural messenger-like conversations in English created by linguists with manually annotated summaries. DialogSum~\citep{chen2021dialogsum} is a recently released dataset for a more challenging task with a lower compression ratio. It contains multi-turn dialogues of real-life scenarios collected from three dialogue corpora. The data statistics are in Table~\ref{tab:dataset_stat}.

%The following models are adopted as baselines: (1) PointerGenerator~\citep{see2017get}; (2) DynamicConv~\citep{wu2019pay}; (3) Transformer~\citep{vaswani2017attention}; (4) DialoGPT~\citep{zhang2019dialogpt}; (5) UniLM~\citep{dong2019unified}; (6) PEGASUS~\citep{zhang2020pegasus}; (7) BART-xsum~\citep{lewis2019bart}; (8) D-HGN~\citep{feng2021incorporating}; (9) S-BART~\citep{chen2021structure}; (10) CODS~\citep{wu2021controllable};
We adopt three different types of baselines: (i) generative language models~\citep{see2017get,wu2019pay,vaswani2017attention}; (ii) pre-trained language models~\citep{zhang2019dialogpt,dong2019unified,zhang2020pegasus,lewis2019bart}; (iii) dialogue summarization Models~\citep{feng2021incorporating,chen2021structure,wu2021controllable}. We provide more details in Appendix~\ref{sec:baselines}.

\subsection{Implementation Details}

We employ two automatic evaluation metrics as: (i) ROUGE~\citep{lin2004rouge} scores, including ROUGE-1, ROUGE-2, and ROUGE-L, which compares word-level uni-gram and bi-gram, and the longest common sequence overlap with the gold summary respectively; (ii) BERTScore~\citep{zhang2020bertscore}\footnote{We follow \href{https://github.com/Tiiiger/bert\_score}{https://github.com/Tiiiger/bert\_score} to calculate BERTScore. Note that different tools may result in different BERTScore.}, the recent popular metric for text generation, which computes the contextual similarity score between generated and reference summaries. We report F1 scores for both metrics. For simplicity, we use R-1, R-2, R-L, and B-S to denote ROGUE-1, ROUGE-2, ROUGE-L, and BERTScore (see Appendix~\ref{sec:automatic_metric}).

% We use BART-xsum to initialize our models for training in all experiments. 
Our implementation is based on the Huggingface implementation~\citep{wolf2020transformers} of BART language model. Specifically, we use the weight checkpoint of BART-xsum\footnote{\href{https://huggingface.co/facebook/bart-large-xsum}{https://huggingface.co/facebook/bart-large-xsum}}. We use a maximum input length of 1024 tokens and output length of 100 tokens. Note that the input is either padded or truncated after each utterance and its corresponding commonsense is concatenated during pre-processing. 
We use a learning rate of 3e-6 and a batch size of 32 when fine-tuning our model on both benchmarks. We use linear warm-up over the first 600 steps, apply linear decay and use the Adam optimizer~\citep{kingma2014adam}. In our experiments, we use beam search with beam size of 20. We fine-tune our model on SAMSum for 20 epochs and DialogSum for 25 epochs. All experiments are run on one A100 NVIDIA GPU. More implementation details about commonsense knowledge generation is included in Appendix~\ref{sec:implementation_commonsense}.

%We tried two optimizers: Adam (Kingma & Ba, 2015) with weight decay of 0.01 (as recommended by (Devlin et al., 2019)) and Adamax (Kingma & Ba, 2015) without weight decay; based on val- idation set performance, we choose to fine-tune with Adam when using the BERT weights. The learning rate is initialized to 5e-5 with a warmup of 100 iterations for Bi- and Poly-encoders, and 1000 iterations for the Cross-encoder. The learning rate decays by a factor of 0.4 upon plateau of the loss evaluated on the valid set every half epoch. In Table 3 we show validation performance when fine-tuning various layers of the weights provided by (Devlin et al., 2019), using Adam with decay optimizer. Fine-tuning the entire network is important, with the exception of the word embeddings.

\section{Experimental Results}
%% semantic 적으로 모호한 부분을 분명하게 해준다. 

\subsection{Automatic Evaluation}\label{sec:automatic}

%For the paper decision recommendation task, HabNet achieves the best performance results no matter which kind of embedding is used. This demonstrates the effectiveness of our framework and its generality. To be specific, compared with flat baselines, our framework with GloVe embedding, i.e., HabNet (Glove), performs much better, which demonstrates
%We achieve a new state-of-the-art on both text8 and enwik8 using the small models with BPC of 1.10 and 1.00 on text8 and enwik8 respectively, demonstrating the effectiveness of our model.

\vspace{0.8mm}
\noindent\textbf{Performance}
Table~\ref{tab:automatic_evaluation} presents the performance on SAMSum and DialogSum test sets. SICK++ outperforms all baselines on ROUGE-1, ROUGE-2 and BERTScore by a consistent margin in both datasets. 

\vspace{0.8mm}
\noindent\textbf{Comparison with State-of-the-Art}
We find that pre-trained language models (\eg, DialoGPT, UniLM, PEGASUS, BART-xsum), outperform models that are not pre-trained (\eg, PointerGenerator, DynamicConv, Transformer), confirming the impact of pre-training on abstractive dialogue summarization. Among the pre-trained generative language models examined, PEGASUS and BART-xsum are the most competitive models with ROUGE-1 higher than 50. SICK++ shows improvement on all metrics compared to BART-xsum (\eg, without additional input, commonsense supervision) in both benchmarks showing that our method can be applied in different settings.
%SICK was fine-tuned using BART-xsum as a backbone model, and therefore we reimplemented using the same hyper-parameter configurations from \ref{implementation details}

Among methods that alter the input to seek additional useful information in a dialogue setting, (\eg, D-HGN, SBART, and CODS), CODS achieves better performance over other baselines in SAMSum. However, on DialogSum, a more challenging setting due to higher abstractiveness, CODS is not able to get as much gain of performance compared to other baselines. Meanwhile, SICK++ outperforms all baselines and shows competitive results implying the robustness of our framework.

\vspace{0.8mm}
\noindent\textbf{Commonsense Models}
While SICK++ shows better performance regardless of which commonsense generation model is used, the excelling choice differs depending on the dataset. In SAMSum, SICK++ shows better performance with PARA-COMET than with COMET, however it shows opposite result in DialogSum. We conjecture this due to the characteristic of datasets and commonsense models hold. 
PARA-COMET has an advantage of using parametric memory to consider previous sentences, which may be sensitive in terms of length. Since SAMSum has shorter length of dialogues than DialogSum, the recurrent memory component of PARA-COMET is less likely to forget the previous sentences. We expect to get better performance with the help of commonsense-models that maintains longer memories of sentences/dialogues and leave this as future research. 

\begin{table}[t!]
\begin{center}\small
{\begin{tabular}{l cc cc}
    \toprule
    & \multicolumn{2}{c}{SAMSum} & \multicolumn{2}{c}{DialogSum} \\
    \cmidrule(lr){2-3} \cmidrule(lr){4-5}
    Model & Info. & Cons. & Info. & Cons. \\
    \midrule
    BART-xsum & 3.71 & 3.48 & 3.71 & 3.68 \\
    SICK++ & 3.85 & 3.81 & 3.79 & 3.97 \\
    \midrule
    Gold & 4.00 & 3.96 & 4.03 & 4.21\\
    \bottomrule
\end{tabular}}
\end{center}
\caption{Human evaluation on SAMSum and DialogSum datasets. Info. and Cons. denotes informativeness and factual consistency respectively.}
\label{tab:human_evaluation}
\end{table}

\begin{comment}
\begin{table}[t!]
\begin{center}\small
{\begin{tabular}{cl cc}
    \toprule
    Dataset & Model & Info. & Cons. \\
    %Flu. & Con. & Rel. & Coh. \\
    \midrule
    \multirow{3}{*}{SAMSum}
    & BART-xsum & 3.71 & 3.52 \\
    & SICK & 3.85 & 3.81 \\
    \cmidrule{2-4}
    & Gold & 4.00 & 3.96 \\
    \midrule
    \multirow{3}{*}{DialogSum}
    & BART-xsum & 3.71 & 3.68 \\
    & SICK & 3.79 & 3.97 \\
    \cmidrule{2-4}
    & Gold & 4.03 & 4.21\\
    \bottomrule
\end{tabular}}
\end{center}
\caption{Human evaluation results on SAMSum and DialogSum datasets. Info. and Cons. denotes informativeness and factual consistency respectively.}
\label{tab:human_evaluation}
\end{table}
\end{comment}
%\begin{comment}
% zeroshot 성능
\begin{table*}[t!]
\begin{center}\small
{\begin{tabular}{l cccc cccc}
    \toprule
    & \multicolumn{4}{c}{SAMSum} & \multicolumn{4}{c}{DialogSum} \\
    \cmidrule(lr){2-5} \cmidrule(lr){6-9}
    Model & R-1 & R-2 & R-L & B-S & R-1 & R-2 & R-L & B-S  \\
    \midrule
    BART-xsum & 20.83 & 4.28 & 15.28 & 46.59 & 17.40 & \textbf{4.16} & 13.80 & 42.97 \\
    SICK & \textbf{23.12} & \textbf{5.09} & \textbf{17.45} & \textbf{47.69} &\textbf{ 18.32} & 3.80 & \textbf{14.98} & \textbf{43.97} \\

    \bottomrule
\end{tabular}}

\end{center}
\caption{Zero-shot evaluation on SAMSum and DialogSum test set.}
\label{tab:zero_shot_samsum_and_dialogsum}
\end{table*}

\subsection{Human Evaluation}\label{sec:humaneval}

We conduct human evaluation to verify the quality of the generated summaries. 
We randomly sample 50 dialogues from test sets of SAMSum and DialogSum, respectively. Annotators were asked to score the quality of a set of summaries from BART-xsum, SICK++, and ground-truth using a Likert scale from 1 (worst) to 5 (best) in terms of \textbf{informativeness} (\ie, covers adequate information) and \textbf{factual consistency} (\ie, consistent with the original input). Each summary was evaluated by three different annotators. Also, the win-loss ratio, which is not biased by subjectivity, is 51.33 (informativeness) and 54.16 (factual consistency), which is consistent to the observations made from the absolute scores.
% The Intra-class correlation is 0.000, indicating moderate agreement~\citep{koo2016aguideline}.

In Table~\ref{tab:human_evaluation}, human annotated summaries receive the best scores on all dimensions. SICK++ gets better scores than BART-xsum for informativeness, which matches the results of ROUGE scores in Section~\ref{sec:automatic}. 
Neural abstractive models often suffer from hallucinations that affect their reliability~\citep{zhao2020reducing}. SICK++ also produces more consistent summaries even though factual consistency is not explicitly modeled. We assume that incorporating commonsense knowledge helps the model recognize the hidden meanings and better understand the dialogue, resulting in fewer factual errors without improper reasoning over conversational flow.

% This might because that the incorporation of structured information such as relations helped sbart to recognize the salinet parts in conversations, and thus improve the succinctness and informativenes sover bart-xsum.

%As shown in Table 7, human annotated summaries receive the best scores from all dimensions. UNILMV2BASE has steadily better scores than Transformer, but lower than human. Model-generated summaries have the highest scores on Fluency, while lowest on Consistency. It suggests that although model-generated summaries are grammatical and fluent, they still contain factual errors.

%On both CNN and Reddit, the plausibility model’s deletions are highly gram- matical, and we also see evidence that the plau- sibility model makes more semantically-informed deletions to maintain factuality, especially on CNN

%Factuality performance is lower on Reddit, but incorporating the plausibility model on top of the compression rules results in a 6% gain in precision. There is still, however, a large gap between factual- ity in this setting and factuality on CNN, which we suspect is because Reddit summaries are different in style and structure than CNN summaries: they largely consist of short event narratives (Kim et al., 2019), and so annotators may disagree on the de- gree to which deleting spans such as subordinate clauses impact the meaning of the events described.
\section{Analysis}

To evaluate the effectiveness of our method, we address the following research questions to guide our experiments:
\begin{itemize}
\item \textbf{RQ1}: Does commonsense help summarizing dialogues?
\item \textbf{RQ2}: Is our method worth using in terms of efficiency despite the extra effort? 
\item \textbf{RQ3}: Does \textit{commonsense supervision} lead SICK++ to inject commonsense knowledge?
\end{itemize}

\subsection{RQ1: Commonsense Applicability}\label{sec:rq1}

We experiment in a zero-shot setting to examine how commonsense knowledge solely affects dialogue summarization . While there exist many factors that could affect performance besides commonsense during training (\eg, hyperparameter configurations), in a zero-shot setting, we can directly compare when commonsense is given and not. We evaluate BART-xsum and SICK on the SAMSum and DialogSum test sets. Note that we use SICK (\ie, only provided input commonsense) instead of SICK++ for zero-shot evaluation, since we cannot access ground-truth summary to generate target commonsense inferences \calZ. %so it is also unable to utilize target commonsense.%For post-processing, we deleted summaries that copy the dialogue identically to test the abstractiveness of the summary.

Table~\ref{tab:zero_shot_samsum_and_dialogsum} presents zero-shot evaluation results on SAMSum and DialogSum respectively.  
We find that SICK outperforms BART-xsum, where the performance gain comes from additional commonsense. Since the only difference between BART-xsum and SICK is the input commonsense, providing extra commonsense for each utterance as Equation~\ref{eq:input_format} helps the model generate more accurate and semantically informative summaries.
This also supports the idea that commonsense is essential in resolving the discrepancy between dialogues and documents.
% the performance gain indicates the effectiveness of commonsense knowlege

% We study the influence of text length on the model performance and computational cost
%For post-processing, we deleted summaries that copy the dialogue identically to test the abstractiveness of the summary. 

%%% 굳이 데이터 추가 하는 비용을 감수하면서도 할만한 이유 

%To verify whether commonsense knowledge can help our model better generalize to the new domain, 

%In the zero-shot environment, we define SICK$_{zero}$ as using Bart-large-xsum weights as a backbone and only providing additional commonsense input, not target output. This is due to the situation in which we cannot provide additional signal via training the model in a zero shot setting.

%%% zero shot 
% Based on the results from table~~~ we observe that SICK has a comparative advantage when the number of training examples are smaller.
% %%% why are we doing this 
% We train SICK in zero and low resource settings to study the 
%%% so how are we going to do this 
%%% what are the results (zero)

% Based on the results from table-to-text (§6.1) and summarization (§6.2), we observe that prefix- tuning has a comparative advantage when the num- ber of training examples is smaller. To explore the low-data setting more systematically, we sub- sample the full dataset (E2E for table-to-text and XSUM for summarization) to obtain small datasets of size {50, 100, 200, 500}

% To examine the benefit of incorporating commonsense knowledge in the dialogue summarization task, we directly evaluate BART-xsum and SICK$_{zero}$ on the SAMSum and DialogSum test sets in zero-shot manner. 

\subsection{RQ2: Data Efficiency}\label{sec:rq2}
\begin{figure*}[t!]
\centering
\includegraphics[width=0.98\linewidth]{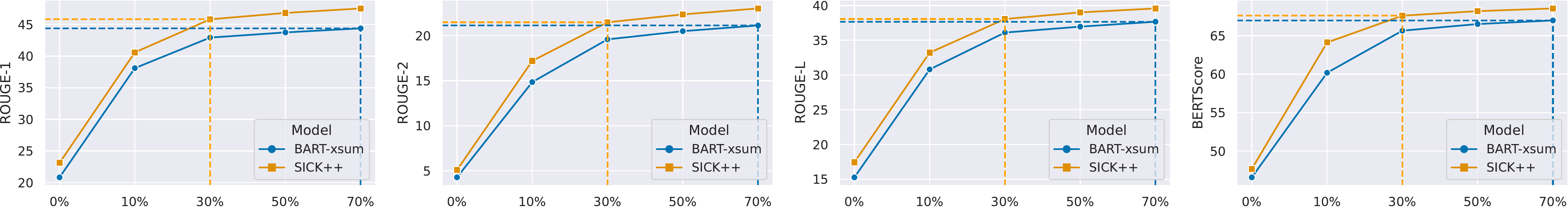}
\caption{Performance of BART-xsum and SICK++ on SAMSum by varying the size of training data. We use ${\textsc{BART}}_{\textsc{BASE}}$ for both of them. Details are shown in Appendix.}
\label{fig:low_resoure}
\end{figure*}
%% 세팅 다시 돌려야 할 수도 있음!
%% 디자인 별로임
% rouge 하나랑 bs 하나 이렇게만 보여주는 건?

% Some may question the performance gain of SICK++ simply
Generating commonsense inferences requires irresistible effort, further described in Appendix~\ref{sec:implementation_commonsense}. Our approach has limitations in terms of time efficiency. However, we find that our method is helpful in situations where data is insufficient, meaning there is a trade-off (\textit{time vs data efficiency}). 

We hypothesize that due to providing additional knowledge and \textit{commonsense supervision}, SICK++ can show comparable performance even if only a small amount of training data is available (\ie, data efficiency). As shown in Figure~\ref{fig:low_resoure}, with only 30\% of training data, SICK++ shows better performance than BART-xsum trained with 70\% of training data. Furthermore, SICK++ consistently outperforms BART-xsum regardless of training data size, proving the robustness of SICK++. The performance gap between SICK++ and BART-xsum can be viewed as a consequence of the leveraged commonsense, based on the fact that BART-xsum is the base architecture of SICK++.

\subsection{RQ3: Effect of \textit{commonsense supervision} on Injecting Commonsense Knowledge}\label{sec:rq3}
\begin{figure}[t!]
\centering
\includegraphics[width=0.9\linewidth]{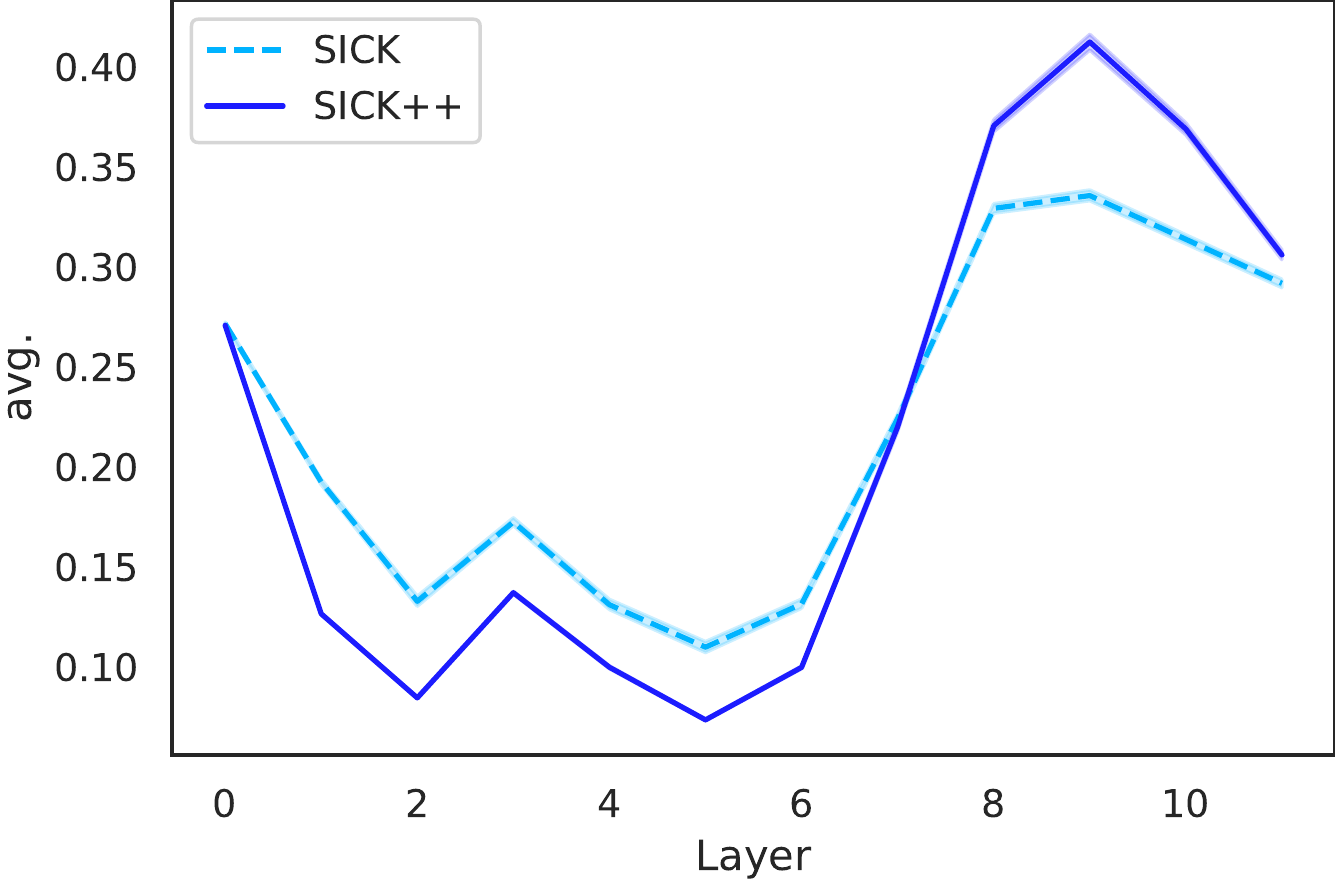}
\caption{Attention visualization of SICK/SICK++. Each point of the line corresponds to the average attention a particular SICK encoder attention head puts towards commonsense inferences.}
\label{fig:attention}
\end{figure}

We observe that SICK++ shows better performance than SICK, as we show in Table~\ref{tab:automatic_evaluation}, but the reason for the performance improvement is somewhat unclear. To analyze the role of \textit{commonsense supervision}, we now take a look at how the dual decoder setting impacts commonsense utilization of the encoder, the difference between SICK and SICK++. 

Attention weights can be viewed as governing how ``important'' every other token is when producing the next representation for the current token~\citep{clark2019does}. We conduct a experiment of measuring the averaged attention value of the commonsense inferences compared to utterances using validation sets of DialogSum, which is more abstractive (\ie, more challenging to comprehend) compared to SAMSum.%For this observation, we compare SICK++ and SICK, which are identical except for the additional decoder $\mathbb{D}_{cs}$ of SICK++. 

%When constructing the encoder input of SICK/SICK++, <I> token is used to provide additional commonsense for each corresponding utterance, functioning as an indicator for commonsense. Following the setting of \citet{clark2019does}, we compute the average attention values of <I> tokens across the attention heads on each layer to analyze attention patterns. Since <I> token is always set in front of commonsense inferences, it can indirectly be viewed as an aggregated commonsense representation such as how [CLS] token represents a sentence~\citep{rogers2020primer}.

The results are illustrated in Figure~\ref{fig:attention}. \citet{rogers2020primer} mentioned that final layers of language models are most task-specific, and we observe that SICK++ has marginally higher attention values. We conjecture this is due to the supervision provided by generating $\calZ$ instead of relying on distant supervision, meaning that our goal of enforcing the model to use commonsense inferences is successful. SICK++ enforces the encoder to fuse the two different modalities (\eg, utterances, commonsense inferences). Meanwhile in lower and middle layers, SICK++'s attention values tend to be lower than SICK. One possible reason is that lower layers tend to look at syntactic and word-level information~\citep{rogers2020primer}, whereas the commonsense inferences generated by COMET or PARA-COMET is only meaningful when understood conceptually.

\section{Conclusion}

%To advance the abstractive dialogue summarization, we propose SICK framework where commonsense knowledge from dialogues and summaries is injected to the model for . We conduct a set of experiments including automatic and human evaluation on benchmarks, and demonstrate that SICK outperforms State-of-the-art baselines and is also preferred by human judges. We further analyze diverse commonsense knowledge models and relation types, and explore the effect of using commonsense into dialogue summarization via attention analysis.

% In this work, we demonstrated that dense retrieval
% can outperform and potentially replace the traditional sparse retrieval component in open-domain
% question answering. While a simple dual-encoder
% approach can be made to work surprisingly well,
% we showed that there are some critical ingredients
% to training a dense retriever successfully. Moreover,
% our empirical analysis and ablation studies indicate
% that more complex model frameworks or similarity
% functions do not necessarily provide additional values. As a result of improved retrieval performance,
% we obtained new state-of-the-art results on multiple
% open-domain question answering benchmarks.

% missing gap can be filled with commonsense knowledge
% 
In this work, we propose SICK and SICK++ framework in order to resolve the two key challenges: i)\; \textit{filling in the gap} in dialogues;  ii)\; injecting commonsense knowledge into a model. We show that the difficulties in dialogues are resolved with commonsense knowledge and demonstrated that our framework can successfully inject commonsense knowledge. As a result of injected commonsense knowledge, we obtain competitive results on SAMSum and DialogSum benchmarks.

% In this work, we demonstrated that our framework SICK and SICK++ can outperform

% In this paper, we propose to utilize commonsense knowledge to resolve the difficulties that lie within dialogue summarization. We introduce SICK, a framework that uses commonsense knowledge as additional input and target supervision instead of solely relying on the input dialogue. Our experimental results show that our method outperforms baselines on SAMSum and DialogSum and has the effectiveness to inject commonsense knowledge to generate a more informative and fluent summary.

\section*{Acknowledgements}
The authors would like to thank the anonymous reviewers for their helpful comments and suggestions. This work was partly supported by Institute of Information \& communications Technology Planning \& Evaluation (IITP) grant funded by the Korea government(MSIT) (No. 2020-0-01361, Artificial Intelligence Graduate School Program (Yonsei University)), (No.2021-0-02068, Artificial Intelligence Innovation Hub), and (No. 2022-0-00077, AI Technology Development for Commonsense Extraction, Reasoning, and Inference from Heterogeneous Data). Jinyoung Yeo is a corresponding author.

\bibliography{acl2022}
\bibliographystyle{acl_natbib}

\appendix

\section{Baselines}\label{sec:baselines}
\vspace{0.8mm}
\noindent\textbf{Generative Language Models}
\begin{itemize}

    \item \textbf{PointerGenerator}~\citep{see2017get} is a  RNN-based method designed for text summarization that deploys copy-attention mechanism.
    \item \textbf{DynamicConv}~\citep{wu2019pay} is a lightweight convolutional model that can perform competitively to self-attention.
    \item \textbf{Transformer}~\citep{vaswani2017attention} is a random-initialized (\ie, not pre-trained) encoder-decoder architecture with self attention and multi-head attention.
\end{itemize}
\vspace{0.8mm}
\noindent\textbf{Pre-trained Generative Language Models}
\begin{itemize}
    \item \textbf{DialoGPT}~\citep{zhang2019dialogpt} is a GPT-2 model pre-trained on open-domain Reddit data designed for response generation.
    \item \textbf{UniLM}~\citep{dong2019unified} is a unified language model which can be used for both natural language understanding and generation tasks by pre-trained using three types of language modeling tasks: unidirectional, bidirectional, and sequence-to-sequence prediction on English Wikipedia and BookCorpus.
    \item \textbf{PEGASUS}~\citep{zhang2020pegasus} is a model specifically designed for summarization tasks where it is pre-trained with an gap-sentence objective. Important sentences are masked from input and is trained to generate the missing parts, similar to an extractive summary approach.
    \item \textbf{BART}~\citep{lewis2019bart} is trained by corrupting text with an arbitrary noising function and learning to reconstruct the original text.
    \item \textbf{BART-xsum}\footnote{\href {https://huggingface.co/facebook/bart-large-xsum}{https://huggingface.co/facebook/bart-large-xsum}} denotes a BART~\citep{lewis2019bart} model fine-tuned on XSUM~\citep{narayan2018xsum} dataset.
\end{itemize}
\vspace{0.8mm}
\noindent\textbf{Dialogue Summarization Models}
\begin{itemize}
    \item \textbf{CODS}~\citep{wu2021controllable} finds key phrases, and generates length-controllable summary from the key phrases.
    \item \textbf{D-HGN}~\citep{wu2021controllable} incorporated commonsense knowledge from ConceptNet~\citep{speer2017conceptnet} for dialogue summarization.
    \item \textbf{S-BART}~\citep{chen2021structure} incorporated discourse relations between utterances, and the connections between speakers and actions within utterances to generate abstractive conversation summarization.
\end{itemize}

\section{Implementation Details of Commonsense Generation}\label{sec:implementation_commonsense}

To generate commonsense, we use COMET and PARA-COMET. Each commonsense model has different choices in terms of model architecture. For COMET, we use BART version among several available versions. Publicly available checkpoints were used for both COMET\footnote{\href{https://github.com/allenai/comet-atomic-2020}{https://github.com/allenai/comet-atomic-2020}} and PARA-COMET\footnote{\href{https://github.com/skgabriel/paracomet}{https://github.com/skgabriel/paracomet}}.GPT-2 version was used for PARA-COMET.For inference, we use beam search with beam size 5 and 10 for each COMET and PARA-COMET, the default setting provided in the public repository. All this procedure is done on one GeForce RTX 3090 GPU. 

To investigate the overhead, we measure the time required to generate commonsense inferences in SAMSum. SAMSum, consisted of 14732 samples within the train subset, took 18.3 hours to generate all the needed commonsense inferences. In other words , it took about 4.4719 seconds per dialogue to generate the commonsense.Note that SAMSum has an average of 11.2 turns per dialogue, so that this number could vary depending on how long the given dialogue is.

\section{Automatic Evaluation Metrics}\label{sec:automatic_metric}

The following metrics are used for the evaluation of baselines and our models:
\begin{itemize}
    \item \textbf{ROUGE} measures the number of overlapping textual units between generated summary and a set of reference summaries. %We report Specifically, ROUGE-1 [[]] , ROUGE-2 [[]] , and ROUGE-L .
    \item \textbf{BERTScore} computes the similarity scores by aligning generated and reference summaries on a token-level based on the output of the BERT-based model. Token alignments are computed greedily with the objective of maximizing the cosine similarity between contextualized token embeddings. We report the F1 score.
\end{itemize}

\section{Human Evaluation Metrics}\label{sec:human_metric}

In general, the gold-standard method for evaluating text generation is still human evaluation, where human annotators assess the quality of generated texts. We adopt the following human evaluation metrics:
%This evaluation can be done from perspectives, and we list a few common varieties below (all are investigated in §4):
\begin{itemize}
    \item \textbf{Informativeness}: How well does the generated summary captures the key ideas of the source dialogue?
    \item \textbf{Factual Consistency}: How consistent is the generated summary with respect to the source dialogue? Does the generated summary contain only statements entailed by the source dialogue?
    %Whether the generated hypothesis contains only statements entailed by the source text.
    %\item \textbf{Conciseness} : does not contain redundant information.
    %\item Factuality (FAC): Whether the generated hypothesis contains only statements entailed by the source text.
    %\item Fluency (FLU): Whether the text has no formatting problems, capitalization errors or obviously ungrammatical sentences (\eg, fragments, missing components) that make the text difficult to read.
    %\item Coherence (COH): Whether the text builds from sentence to sentence to a coherent body of information about a topic.
    %\item Semantic Coverage (COV): How many semantic content units from reference texts are covered by the generated hypothesis.
    % \item Adequacy (ADE): Whether the output conveys the same meaning as the input sentence, and none of the message is lost, added, or distorted.
    % \item Relevancy (REL): How consistent the generated summary is with respect to the source text.
\end{itemize}
%Most existing evaluation metrics were designed to cover a small subset of these perspectives. For example, BLEU aims to capture the adequacy and fluency of translations, while ROUGE was designed to match the semantic coverage metric.
%% fluency, consistency, relevance, coherence, informative

%Qualitative/Human Evaluation We also con- ducted human annotations to evaluate the extracted dialogue summaries, in addition to ROUGE scores. Similar to Gliwa et al. (2019), we asked human annotators on Amazon Mechanical Turk 5 to rate each summary (200 randomly sampled summaries in total) on the scale of [-2, 0, 2], where -2 means that a summary was poor, extracted irrelevant in- formation or did not make sense at all, 2 means it was understandable and gave a concise overview of the text, and 0 refers to that the summary only ex- tracted only a part of relevant information, or made some mistakes. The score for each summary was averaged among three different annotators. The Intra-class Correlation was 0.583, indicating mod- erate agreement (Koo and Li, 2016).

\section{Commonsense Selection Methods}\label{sec:roberta}

\begin{table*}[t!]
\begin{center}\small
{\begin{tabular}{ll cccc cccc}
    \toprule
    & & \multicolumn{4}{c}{SAMSum} & \multicolumn{4}{c}{DialogSum} \\
    \cmidrule(lr){3-6} \cmidrule(lr){7-10}
    Generation Model & Selection Model & R-1 & R-2 & R-L & B-S & R-1 & R-2 & R-L & B-S \\
    \midrule
    \multirow{3}{*}{COMET}
    & Random & 53.04 & 27.17 & 48.49 & 71.34 & 46.05 & 20.46 & 40.61 & 70.84\\
    & NLI-based & 53.21 & 28.02 & 48.85 & 71.53 & 45.26 & 19.94 & 40.04 & 70.54\\
    & Similarity-based & \textbf{53.24} & \textbf{28.10} & \textbf{48.90} & \textbf{71.71} & \textbf{46.31} & \textbf{20.95} & \textbf{41.10} & \textbf{71.71}\\
    \midrule
    \multirow{3}{*}{PARA-COMET}
    & Random & 52.95 & 27.62 & 48.51 & 71.45 & 45.59 & 20.16 & 40.23 & 70.65\\
    & NLI-based & 52.99 & 28.22 & 48.61 & 71.69 & 45.14 & 20.01 & 39.98 & 70.99\\
    & Similarity-based & \textbf{53.73} & \textbf{28.81} & \textbf{49.50} & \textbf{71.92} & \textbf{46.20} & \textbf{20.39} & \textbf{40.83} & \textbf{71.32} \\
    \bottomrule
\end{tabular}
}
\end{center}
\caption{Performance of SICK++ by varying the commonsense related methods.}
\label{tab:selection}
\end{table*}

\begin{comment}
\begin{table}[t!]
\begin{center}\small
{\begin{tabular}{l cccc}
    \toprule
    Method & R-1 & R-2 & R-L & B-S \\
    \midrule
    \multicolumn{4}{l}{\textbf{Using COMET}} \\
    Random & 53.04 & 27.17 & 48.49 & 71.34\\
    NLI & 53.21 & 28.02 & 48.85 & 71.53\\
    Similarity & 53.24 & 28.10 & 48.90 & 71.71\\
    \midrule
    \multicolumn{4}{l}{\textbf{Using PARA-COMET}} \\
    Random & 52.95 & 27.62 & 48.51 & 71.45\\
    NLI \\
    Similarity & 53.73 & 28.81 & 49.50 & 71.92\\
    \bottomrule
\end{tabular}
}
\end{center}
\caption{Performance of SICK on SAMSum test set by varying the commonsense related methods.}
\label{tab:selection_samsum}
\end{table}

\begin{table}[t!]
\begin{center}\small
{\begin{tabular}{l cccc}
    \toprule
    Method & R-1 & R-2 & R-L & B-S \\
    \midrule
    \multicolumn{4}{l}{\textbf{Using COMET}} \\
    Random \\
    NLI-based \\
    Similarity-based & 46.26 & 20.95 & 41.05 & 71.30\\
    \midrule
    \multicolumn{4}{l}{\textbf{Using PARA-COMET}} \\
    Random \\
    NLI-based \\
    Similarity-based & 45.71 & 20.55 & 40.59 & 71.04\\
    \bottomrule
\end{tabular}}
\end{center}
%\caption{Comparison of several commonsense selection methods on DialogSum test set.}
\caption{Performance of SICK on DialogSum test set by varying the commonsense related methods.}
\label{tab:selection_dialogsum}
\end{table}
\end{comment}

%\vspace{0.8mm}
%\noindent\textbf{Commonsense Selection} 

% 병우님이  NLI 로 selection하는게 갑자기 왜 튀어나왔는지 어색하다는 의견을 주셨음.
We consider two different methods in addition to our similarity-based method to filter commonsense inferences : (i) Random : any random commonsense inferences from 25 possible candidates are chosen for each utterance; (ii) NLI-based : deploy a pre-trained language model that is fine-tuned on a natural language inference (NLI) task, to determine whether a commonsense inference does not contradict with the utterance/sentence.

We use random selection method as a baseline to compare whether filtering helps gain additional performance. 

NLI based method is also used by previous works~\citep{gabriel2021paragraph, west2021symbolic} to measure the quality of commonsense inferences. Given a pair of $\{u_i, c_i^{r}\}$ or $\{y_i, z_i^{r}\}$, we acquire the probability of \textsc{Entail} and \textsc{Contradict}. Then we measure the score as:
\begin{equation}
\begin{aligned}
    &\text{NLI Score}(u_i, c_i^{r}) = \\
    &P(\textsc{Entailment}) - P(\textsc{Contradict}) \\
\end{aligned}
\end{equation}
\begin{equation}
\begin{aligned}
    &\text{NLI Score}(y_i, z_i^{r}) = \\
    &P(\textsc{Entailment}) - P(\textsc{Contradict}) \\
\end{aligned}
\end{equation}
where the commonsense inference with the highest NLI Score is selected. As a result, we obtain the input commonsense $\calC = \{ c_1, c_2, ..., c_n \}$ aligned with dialogue $\calD$ for additional context and the target commonsense $\calZ =  \{ z_1, z_2, ..., z_m \}$ aligned with summary $\calY$ for additional supervision.

%Diversity of generated commonsense might act as a double-edged-sword, so it should be controlled. To control the diversity we use 
% random, NLI, similarity를 선정한 이유에 대해서 언급해서 설득해줘야 함 
% heatmap 실험을 통해서 결과의 근거를 설명하는게 좋을 듯. 가장 best는 overlap 점수가 random, nli selection method보다 sbert를 이용하는 것이 가장 잘 나온다면 논리가 맞을 것 같음.

For NLI-based selection, we use RoBERTa-Large~\citep{liu2019roberta} which is fine-tuned on MNLI~\citep{williams2018mnli} to score commonsense candidates. Note that we do not have any label telling which commonsense inference is most plausible to be chosen when given an utterance, therefore, we measure the NLI scores in a zero-shot manner.

As shown in Table~\ref{tab:selection}, using the similarity-based selection method consistently outperforms other methods, regardless of the type of commonsense knowledge model. Since NLI-based method is more intuitive compared to similarity-based methods, and was used in previous works, one might ask why NLI-based method does not show good performance. We conjecture this due to the complexity of each task. Measuring the relation of inclusion is more complex in nature compared to simply measuring the semantic similarity. Our methodology uses a zero-shot setting, therefore it is harder to reach the standards without supervision. The outperforming choice of commonsense selection method could differ when trained with labeled data, and we leave this to future work.

Also, one might conjecturfe that using the top-1 selected commonsense inference with the similarity-based method is a copy of the utterance, resulting in inferences with similarity value 1.0 only selected. However, we found that the mean value of the top-1 commonsense inferences are 0.535799, and standard deviation 0.176364. This shows that the commonsense inferences isn't a copy except for a few bad cases. Considering both diversity and quality is important, and we also leave this to future work.

\section{Choose of Commonsense Relations from COMET}
In prior work such as \citet{chakrabarty2022s} and \citet{li2021past}, it is conventional to selectively use a subset of the COMET relations, depending on the characteristics of a target domain and task. In our work, the social-interaction relations such as \textit{xIntent} and \textit{xWant} are most preferred with the best performance as they are strongly relevant to human-human interaction in dialogue.

\begin{table*}[t!]
\begin{center}\small
{\begin{tabular}{l p{0.6\textwidth} p{0.3\textwidth}}
    \toprule
    \multicolumn{2}{l}{\textbf{Dialogue}} & \textbf{Commonsense} \\
    Frank: & Son, will you come home this weekend? & Frank has to go to work..\\
    Son: & Not sure yet. & son is not sure yet.\\
    Son: & Something happened? & Person asks to son what happened.\\
    Frank: & Of course not. & Frank doesn't want to be rude.\\
    Frank: & Your mother miss you. & your mother misses you..\\
    Son: & I miss her too. & son misses his mother.\\
    Frank: & So will you come? & Frank is too shy to ask..\\
    Son: & I will try. & son will try\\
    Frank: & Good, I will tell your mother that you will come. & son will come.\\
    Son: & Oh, dad.. ok I will come. & Person asks if he can come.\\
    \multicolumn{2}{l}{\textbf{Gold Summary}} \\
    \multicolumn{2}{l}{Son is coming to see his parents’ this weekend.} \\
    \multicolumn{2}{l}{\textbf{BART-xsum}} \\
    \multicolumn{2}{l}{Son will come home this weekend.} \\
    \multicolumn{2}{l}{\textbf{SICK}} \\ 
    \multicolumn{2}{l}{Son will come home this weekend. He misses his mother.} \\
    
    \midrule
    
    Julie: & \text{<}file photo\text{>} & Julie sent a photo. \\
    Emily: & <3 Julie Love, I'm sending tons of kisses ;*;*;* & to show love. \\
    Emily: & \text{<}emoji\text{>} & Emily sent a photo. \\
    Julie: & Merry Christmas and a lovely mood throughout the whole year, darling. & Julie gives a hug \\
    Emily: & Thank you, for you too <3 & Person is thanked.\\
    Julie: & Thanks:* & Julie gets a hug. \\
    Julie: & \text{<}file photo\text{>} \text{<}file photo\text{>} & Julie sent a photo. \\
    \midrule
    \multicolumn{3}{l}{\textbf{Gold}} \\ 
    \multicolumn{3}{l}{Emily and Julie wish Merry Christmas to each other.} \\
    \multicolumn{3}{l}{\textbf{SICK++}} \\ 
    \multicolumn{3}{l}{Julie and Emily are exchanging Christmas greeting.}\\
    \multicolumn{3}{l}{\textbf{BART-xsum}} \\ 
    \multicolumn{3}{l}{Julie sends Emily tons of kisses.} \\
    
    \midrule

    Stewart: & Can you believe he even said that about the forests & the forest to be healthy. \\
    Stewart: & Raking? Really? & to think about the situation. \\
    Shari: & Yes... I can believe that this is an ignorant man... &  Shari doesn't want to be ignorant. \\
    Shari: & He proves it daily.. This is just one more example! & Shari wants to be helpful.  \\
    Stewart: & He just has no clue... & he has no clue...  \\
    Stewart: & I mean, there are so many people dead and all he can think to do is\\
    & criticize the forestry department? With a totally inappropriate suggestion? & Shari thinks it's inappropriate. \\
    Stewart: & I can't wait to vote for anyone else but him... & to vote for someone else. \\
    Shari: & I know what you mean.. Half my friends voted for him & \\
    & just to see what would happen! Well, guess what? & Shari votes for him \\
    Stewart: & Yeah, we couldn't go another 4 years with a Democrat.. & They want to get rid of him.\\
    & ... & \\
    \midrule
    \multicolumn{3}{l}{\textbf{Gold}} \\ 
    \multicolumn{3}{l}{Stewart and Shari find the current president ignorant and incompetent. They hope he gets voted out. Stewart is going to see } \\
    \multicolumn{3}{l}{what possibilities there are of volunteering in the upcoming elections.} \\
    \multicolumn{3}{l}{\textbf{SICK++}} \\ 
    \multicolumn{3}{l}{Stewart and Shari don't like the fact that the current president raked the forests. They think he's an ignorant man. Shari and}\\
    \multicolumn{3}{l}{Stewart don't want to vote for him, but they have to make the best of it now.}\\
    \multicolumn{3}{l}{\textbf{BART-xsum}} \\ 
    \multicolumn{3}{l}{Stewart and Shari don't like the way the president is behaving. They are going to vote for anyone else but him.} \\
    
    \bottomrule
\end{tabular}}
\end{center}
\caption{Successful examples of generated summaries with SICK from DialogSum.}
\label{tab:success_example}
\end{table*}

\begin{table*}[t!]
\begin{center}\small
{\begin{tabular}{l p{0.6\textwidth} p{0.3\textwidth}}
    \toprule
    \multicolumn{2}{l}{\textbf{Dialogue}} & \textbf{Commonsense} \\
    Person1: & Are you familiar with American-styled accounting? & Person1 asks PersonY if they are familiar with accounting.  \\
    Person2: & I am afraid not. & Person2 is too afraid. \\
    Person2: & I haven't worked in an American company so far. & Person2 is too young to work.\\
    Person1: & What are the most fundamental concepts underlying the accounting process? &  to learn about accounting. \\
    Person2: & The first is accounting entity, and the second is going concern. & Person2 is not qualified. \\
    Person2: & The third is measuring unit. &  Person2 doesn't know how to measure. \\
    Person2: & The fourth is accounting period, and the fifth is objectivity. & Person2 has to be objective. \\
    \midrule
    \multicolumn{3}{l}{\textbf{Gold}} \\ 
    \multicolumn{3}{l}{Person2 tells Person1 about the fundamental concepts of the accounting process.} \\
    \multicolumn{3}{l}{\textbf{SICK++}} \\ 
    \multicolumn{3}{l}{Person2 tells Person1 the most fundamental concepts underlying the accounting process.}\\
    \multicolumn{3}{l}{\textbf{BART-xsum}} \\ 
    \multicolumn{3}{l}{Person1 asks Person2 about American-styled accounting.} \\
    \midrule
    
    Person1 & Oh, it's getting late. & Person1 has to go to work.. \\
    Person1 & I've got to run. &  to be running. \\
    Person1 & It was nice talking to you, karren. &  Person1 calls back. \\
    Person2 & Thanks, Tim. & to talk to Tim. \\
    Person2 & Nice meeting you, too. & to meet PersonY. \\
    Person1 & I guess we'll see each other around. & Person1 calls PersonY. \\
    Person2 & Yeah, I hope so. & Person2 asks Person2 if they are sure. \\
    Person2 & Well, take it easy. & Person2 has to work. \\
    Person1 & You too. & to talk to PersonY.\\ 
    \midrule
    \multicolumn{3}{l}{\textbf{Gold}} \\ 
    \multicolumn{3}{l}{Tim and Karren say goodbye.} \\
    \multicolumn{3}{l}{\textbf{SICK++}} \\ 
    \multicolumn{3}{l}{Tim and Karren say goodbye to each other.}\\
    \multicolumn{3}{l}{\textbf{BART-xsum}} \\ 
    \multicolumn{3}{l}{Tim and Karren meet each other for the first time.} \\
    
    \midrule

    Person1 & Taxi! &  Person1 calls a taxi. \\
    Person2 & Where to, sir? & Person2 asks for directions. \\
    Person1 & I'd like to go to the railway station please. &  to go to the train. \\
    Person2 & Please hop in. &  PersonY asks PersonY to get in..  \\
    Person1 & Is it a long run to the station? & to go to the station.  \\
    Person2 & It'll take about 20 minutes. & PersonY asks how long it will take.  \\
    Person1 & The streets are heavy with traffic at this time of a day, are they? & the traffic is heavy. \\
    Person2 & Yes, they are. & Person2 doesn't know what they are. \\
    Person1 & Is it the rush hour now? &  Person1 has to go to work. \\
    Person2 & Yes, it is. &  Person2 doesn't know if it is.\\
    Person2 & Are you in a hurry sir? & Person2 asks PersonY to hurry up. \\
    Person1 & No, I'm not. & No, I'm not. \\
    Person1 & Would you please drive slowly and carefully? & Person1 asks Person2 to slow down. \\
    Person2 & Yes, sir. & Person2 is asked a question. \\

    \midrule
    \multicolumn{3}{l}{\textbf{Gold}} \\ 
    \multicolumn{3}{l}{Person1 takes a taxi to the railway station in the rush hour.} \\
    \multicolumn{3}{l}{\textbf{SICK++}} \\ 
    \multicolumn{3}{l}{Person1 takes a taxi to the railway station.}\\
    \multicolumn{3}{l}{\textbf{BART-xsum}} \\ 
    \multicolumn{3}{l}{Person1 calls a taxi to go to the railway station. Person2 tells him it'll take about 20 minutes and drives slowly and carefully.} \\
    
    \bottomrule
\end{tabular}}
\end{center}
\caption{Successful examples of generated summaries with SICK from DialogSum.}
\label{tab:success_example}
\end{table*}

\begin{table*}[t!]
\begin{center}\small
{\begin{tabular}{p{0.2\textwidth}| l l}
    \toprule
    \textbf{Error Type}&\textbf{Dialogue} & \textbf{Commonsense} \\
    \midrule
    \multirow{3}{4em}{\mbox{\textbf{Copying Utterance}}} 
     & \#Person2\#: Have a good day!                              & have a good day.                                   \\    
     & \#Person2\#: Well, take it easy.                           & to take it easy.                                  \\    
     & \#Person1\#: Were you born in Los Angeles?                 & born in Los Angeles.                                  \\

    \midrule
    \multirow{3}{4em}{\mbox{\textbf{Factual Consistency}}} 
    & \#Person2\#: I'm afraid not.                                & Person2 is too afraid.\\
    &\#Person2\#: But I'm not sleepy, darling.                    & Person2 is sleepy. \\
    &\#Person2\#: I haven't worked in an American company so far. & Person2 is too young to work.\\
    
    \midrule
    \multirow{3}{4em}{\mbox{\textbf{Not Informative}}} 
    & \#Person2\#: I'm afraid not.                                & Person2 is too afraid.\\
    &\#Person1\#: No, not much.                                   & Person1 says no. \\
    &\#Person2\#: I've heard this one before.                     & Person2 thinks.\\

    \bottomrule
\end{tabular}}
\end{center}
\caption{Failed examples of generated summaries with SICK.}
\label{tab:fail_example}
\end{table*}

\end{document}